\documentclass[12pt]{article}
\usepackage[utf8]{inputenc}
\usepackage{amsmath,amssymb,amsfonts}
\usepackage{graphicx}
\usepackage{booktabs}
\usepackage{caption}
\usepackage{hyperref}
\usepackage{xcolor}
\usepackage{multirow}
\usepackage{array}
\usepackage{float}

\usepackage[backend=biber, 
            sorting=none,
            maxnames=99]{biblatex}
\nocite{*}

\title{QV May Be Enough: Toward the Essence of Attention in LLMs}
\author{Edward Zhang \footnote{contact with:windyrobin@aliyun.com}}
\date{}
\begin{document}

\maketitle

\begin{abstract}

Starting from first principles and a linguistic perspective centered on part-of-speech (POS) and syntactic analysis, this paper explores and derives the underlying essence of the Query-Key-Value (QKV) mechanism within the Transformer architecture. Based on this theoretical foundation, we provide a unified explanatory framework for the efficacy of contemporary architectures, including MQA, GQA, and MLA, while identifying their inherent trade-offs and potential optimization trajectories. We introduce the \textbf{QV paradigm} and provide empirical evidence for its validity. Building upon this, we propose the \textbf{QV-Ka} (Key-after-value \& ctx) optimization scheme, which is further substantiated through experimental validation. The interpretable theoretical analysis of the QKV mechanism presented in this work establishes a robust foundation for the future evolution of large language model architectures.
\end{abstract}

\section{Introduction}
Since its inception, the Query-Key-Value (QKV)\cite{vaswani2017attention} mechanism has been ubiquitous as the architectural cornerstone of the Transformer model. However, the fundamental motivation behind such a conceptual abstraction—and the precise functional roles of Q, K, and V—remains an open question. While the seminal work interprets these terms through the lens of retrieval systems (i.e., Query, Key, and Value), such definitions align more closely with database management paradigms than with linguistic principles. This prompts a critical inquiry: from the standpoint of natural language processing, what linguistic entities do these components represent, and what is the intrinsic logic governing their internal dynamics?

\section{Background}
In the AGF paper (Zhang \cite{zhang2026agf}), we previously introduced the principles of the Attention mechanism. From the perspective of part-of-speech (POS) and syntactic analysis, consider the sentence: "There is a beautiful girl." Within this structure, "girl" serves as the noun (object), while "beautiful" acts as an adjective modifying "girl." Standard linguistic intuition suggests that adjectives have a natural affinity for specific nouns; for instance, the semantic compatibility between "beautiful" and "girl" is significantly higher than that between "beautiful" and "man."

\begin{figure}[h]
    \centering
    \includegraphics[width=0.6\linewidth]{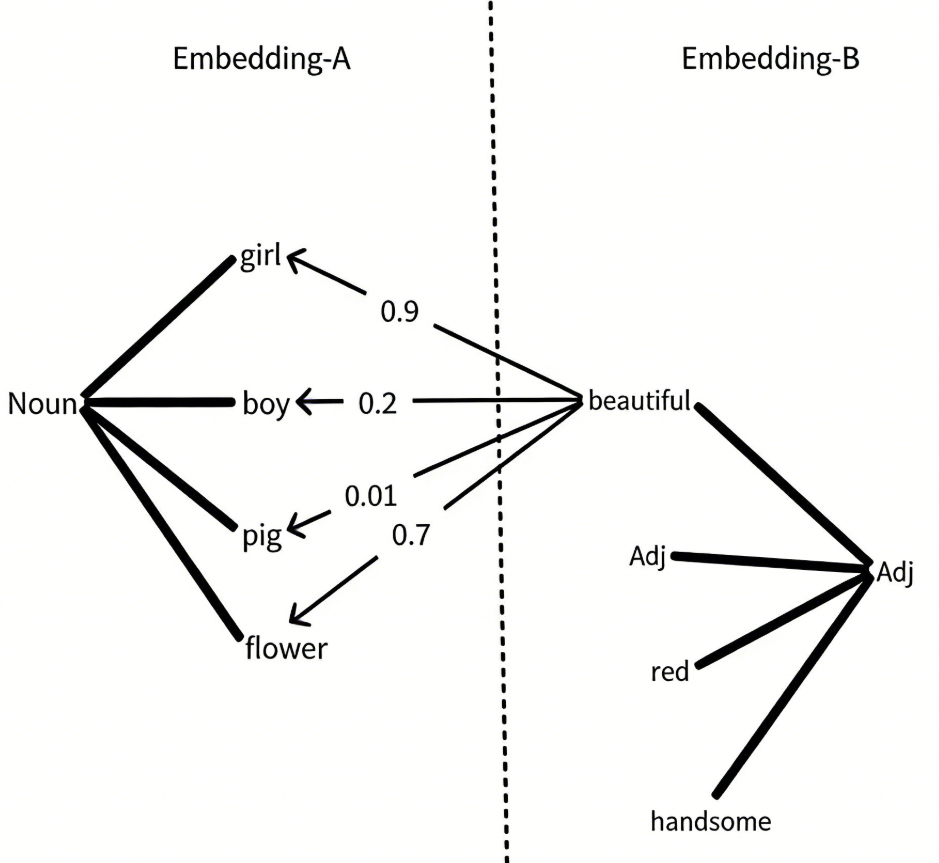}
    \caption{Matching of Different Tokens}
    \label{fig:placeholder}
\end{figure}

Consequently, for any given token embedding, a most-compatible set of related tokens can be identified based on common-sense or statistical heuristics. We define this degree of correlation—or the probability of establishing a modifier-head relationship—as \textit{Affinity} or the \textit{Attention Score}. As established in the AGF framework, this affinity follows a power-law distribution with respect to the relative distance between tokens, fundamentally conforming to the logic of an ``Attention's Gravitational Field.''

\section{QKV Logic}
Next, we will conduct a deeper analysis to demonstrate how the aforementioned Attention logic is implemented. In Attention calculations, $Q, K$, and $V$ are derived from the original Token/Embedding. The core functions of their corresponding transformation matrices $W_q, W_k, W_v$ can be divided into two categories (ignoring positional components for now):

\begin{itemize}
    \item \textbf{Shallow-Composing}: This involves selecting, grouping, or combining previously scattered features. For example, if two dimensions represent ``beautiful'' and ``girl,'' they might be combined into a new temporary feature to express a semantic meaning similar to ``beauty.''
    \item \textbf{Deep-Matching}: (which implicitly includes shallow-composing) This generates pairing or modifying relationships. For instance, when ``girl'' acts as a noun, it often forms a relationship with adjectives like ``young'' or ``beautiful,'' or with other verbs and quantifiers. Common sense tells us this is entirely achievable; this step is the core problem we explored in the AGF paper. Its essence is a probability density issue, conforming to the logic of an ``Attention-Gravitational Field.''
\end{itemize}
\begin{figure}
    \centering
    \includegraphics[width=0.5\linewidth]{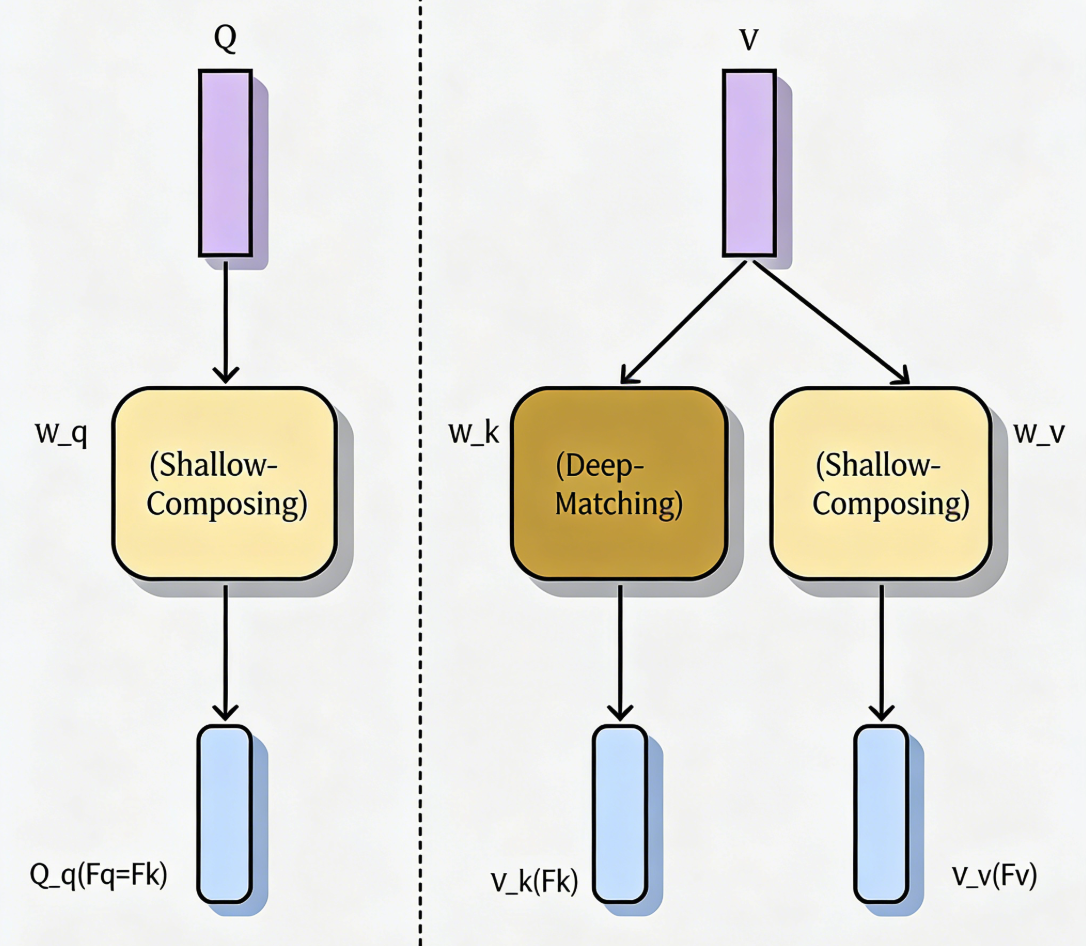}
    \caption{QKV Paradigm}
    \label{fig:placeholder}
\end{figure}

Taking the Transformer's $QKV$ abstraction as an example, if $Q$ comes from $\text{Token-Q}$, and $K, V$ come from $\text{Token-V}$, the resulting semantics can be understood as:

\begin{itemize}
    \item $V_v$: The numerical result of \textit{Shallow-Composing} on $\text{Token-V}$, with its corresponding feature/dimension group denoted as $F_v$.
    \item $V_k$: The result of \textit{Deep-Matching} $\text{Token-V}$ in a certain feature direction $F_k$ (e.g., ``girl'' mapping to the candidates [``beautiful'', ``young'']).
    \item $Q_q$: The projection result of \textit{Shallow-Composing} $\text{Token-Q}$ onto the corresponding feature dimension $F_k$.
\end{itemize}

\noindent \textbf{Note}: The above applies to the scenario where $d_k = d_{model}/h$. In scenarios where $d_k > d_{model}/h$, it can be considered that some dimensions in the feature space $F_v$ undergo ``feature splitting'' within $F_k$ (similar to the logic of ``beauty'' splitting into ``beautiful'' and ``girl''). This effectively improves the ``resolution'' of the Attention correlation calculation, which theoretically enhances model accuracy.

\section{QV Paradigm}

A critical question arises: can the standard $QKV$ logic be simplified? If we aim to eliminate the specific $V_k$ component, the framework transitions as follows:

\begin{itemize}
    \item ${V_v}$: The semantic definition remains consistent with the $QKV$ model, representing the result of \textit{shallow-composing}, with its corresponding feature dimension still denoted as $F_v$.
    \item ${Q_v}$: $\text{Token-Q}$ performs a \textit{deep-matching} calculation along the $F_v$ direction. Logically, this is equivalent to the inverse process of the $V_k$ calculation in the standard $QKV$ model. The resulting mapping would resemble: $(\text{'beautiful'} \rightarrow [\text{'girl', 'flower'}])$.
\end{itemize}

\begin{figure}
    \centering
    \includegraphics[width=0.5\linewidth]{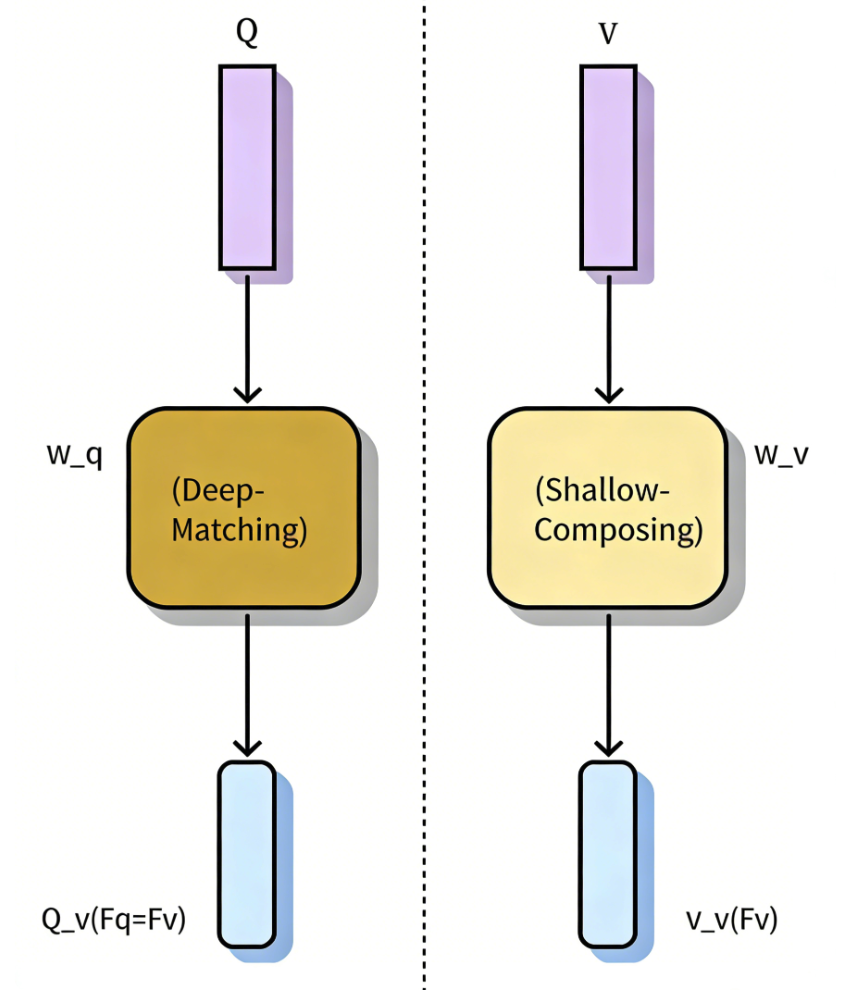}
    \caption{QV Paradigm}
    \label{fig:placeholder}
\end{figure}
In this simplified architecture, both $V_v$ and  $Q_v$ are grounded in the same feature dimension. Their numerical values similarly represent \textit{fact} and \textit{expectation}, respectively. 
\vspace{0.5em}

In other words, the original standard $QKV$ logic essentially follows the form of $Q_qV_kV_v$, 
By simultaneously utilizing $V$ to function as $K$, we derive a simplified mode expressed as:
$Q_vV_v$

\vspace{0.5em}
Consequently, the original Attention calculation formula:
\begin{equation}
    \operatorname{Attention}(Q, K, V) = \operatorname{softmax} \left( \frac{QK^T}{\sqrt{d_k}} \right) V
\end{equation}
is now transformed into:
\begin{equation}
    \operatorname{Attention}(Q, V) = \operatorname{softmax} \left( \frac{QV^T}{\sqrt{d_k}} \right) V
\end{equation}

\section{Experimental Setup}

The experimental methodology in this paper remains consistent with the AGF paper. All experiments are conducted using the \textbf{Vanilla Transformer} as the baseline, specifically employing the \textbf{Transformer-BIG} model architecture.

\begin{table}[htbp]
\centering
\caption{Summary of Experimental Configurations and Hardware Environment.}
\label{tab:experimental_setup}
\begin{tabular}{ll}
\hline
\textbf{Configuration Item} & \textbf{Value / Description} \\ \hline
Base Architecture           & Vanilla Transformer (Transformer-BIG) \\
Dataset                     & WMT\_17 (Translation Task, en-de) \\
Number of Layers            & 3 (Reduced from default 6) \\
Precision                   & FP16 (Half-Precision) Training \\
Hardware                    & Single NVIDIA Tesla V100-PCIE-32GB \\
Training Duration           & $\approx$ 15 hours per run \\
\hline
\end{tabular}
\end{table}

To accelerate the training process, we employed FP16(Half-Precision)training, the number of layers(LayerNum) was set to 3, rather than the default 6.

\vspace{1em}
Training framework  \footnote{ OpenNMT-py: \url{https://github.com/OpenNMT/OpenNMT-py}}  and  AGF module code  \footnote{AGF git: \url{https://github.com/windyrobin/AGF/tree/main}} is avaiable now.

\section{Vanilla-QV Testing and Analysis}

Under the baseline configuration of a \textbf{Vanilla-Transformer} with default \textbf{Sinusoidal Positional Encodings (PE)}, we conducted a comparative analysis between the proposed \textit{QV mode} and the original \textit{QKV mechanism}. The validation accuracy results are summarized in Table~\ref{tab:accuracy_comparison}.

\begin{table}[h]
    \centering
    \caption{Comparison of Validation Accuracy between QV and QKV Modes}
    \label{tab:accuracy_comparison}
    \begin{tabular}{lc}
        \hline
        \textbf{Mode} & \textbf{Valid Accuracy (\%)} \\
        \hline
        QV            & 70.0756 \\
        QKV           & 70.5911 \\
        \hline
    \end{tabular}
\end{table}

As observed, the \textit{QV mode} incurs a performance degradation of approximately \textbf{0.5\%} in accuracy compared to the standard \textit{QKV} approach. 

We hypothesize two primary factors contributing to this 0.5\% loss:

\begin{enumerate}
    \item \textbf{Intrinsic Structural Superiority}: The standard $QKV$ mechanism may inherently possess a more robust representational capacity than the simplified $QV$ mode, naturally leading to higher accuracy.
    \item \textbf{Impact of Positional Encoding (PE) Interference}: In the traditional $QKV$ mechanism, $Q$ and $K$ are injected with PE to compute relative positional relationships, while $V$ remains relatively isolated or less affected by positional interference. However, in the \textit{QV mode}, since $V$ also assumes the role of the original $K$ for positional calculations, it is subjected to a greater degree of interference from the PE. This increased coupling between semantic value and positional information potentially leads to the observed drop in final accuracy.
\end{enumerate}

Based on these observations, we pose a critical research question: \textit{What is the relative magnitude of these two impacts? Specifically, how significant is the interference caused by Positional Encoding (PE)?}

\section{Introduction of AGF and Comparative Analysis}

To eliminate the interference related to positional calculations in the aforementioned comparison, we introduce the \textbf{AGF (Attention's Gravitational Field)} relative positional framework. This approach completely decouples semantic information from positional logic. Furthermore, we apply the standard \textbf{PCM-V} optimization method.
\vspace{0.5em}

The original Attention formulation:
\begin{equation}
    a_{m,n}=\frac{\exp \left( \boldsymbol{q}_{m}^{\top} \boldsymbol{k}_{n} / \sqrt{d} \cdot \text{PosCoeff} \right)}{\sum_{i=1}^{L} \exp \left( \boldsymbol{q}_{m}^{\top} \boldsymbol{k}_{i} / \sqrt{d} \cdot \text{PosCoeff} \right)}, \quad \boldsymbol{o}_{m}=\sum_{n=1}^{L} a_{m, n} \cdot \text{PosCoeff} \cdot \boldsymbol{v}_{n}
\end{equation}
is now transformed into the following $QV$ structure:
\begin{equation}
    a_{m,n}=\frac{\exp \left( \boldsymbol{q}_{m}^{\top} \boldsymbol{v}_{n} / \sqrt{d} \cdot \text{PosCoeff} \right)}{\sum_{i=1}^{L} \exp \left( \boldsymbol{q}_{m}^{\top} \boldsymbol{v}_{i} / \sqrt{d} \cdot \text{PosCoeff} \right)}, \quad \boldsymbol{o}_{m}=\sum_{n=1}^{L} a_{m, n} \cdot \text{PosCoeff} \cdot \boldsymbol{v}_{n}
\end{equation}

The experimental results are summarized in Table~\ref{tab:agf_results}:

\begin{table}[h]
    \centering
    \caption{Comparative Experimental Results with AGF and PCM-V Optimization}
    \label{tab:agf_results}
    \begin{tabular}{llc}
        \hline
        \textbf{Mode} & \textbf{Crafts / Configuration} & \textbf{Valid Accuracy (\%)} \\
        \hline
        QV  & Default (Sinusoidal PE) & 70.0756 \\
        QKV & Default (Sinusoidal PE) & 70.5911 \\
        \hline
        QV  & AGF + PCM-V             & 70.5188 \\
        QKV & AGF + PCM-V             & 70.7800 \\
        \hline
    \end{tabular}
\end{table}

The results clearly demonstrate that after introducing the AGF relative positional calculation, the \textit{QV-AGF} mode substantially matches the performance of the \textit{Vanilla-QKV} baseline. Moreover, the performance gap between \textit{QV} and \textit{QKV} narrows to approximately \textbf{0.26\%}. This indicates that of the original 0.5\% accuracy gap, positional encoding interference accounted for roughly half (approximately 0.2\%--0.3\%).

\vspace{0.5em}
These experimental findings align perfectly with our expectations. The fact that the $QV$ mode can closely approximate the effectiveness of the $QKV$ mode provides robust empirical evidence supporting our hypotheses regarding \textit{Shallow-Composing (SC)}, \textit{Deep-Matching (DM)}, and the underlying logic of $QKV/QV$ abstractions.

\section{Why QKV Outperforms QV}
A pivotal question remains: why does the $QKV$ mechanism consistently achieve higher accuracy than the $QV$ mode? 

\vspace{0.5em}
According to our definition of \textit{Deep-Matching}, the underlying logic involves assessing the degree of correlation or matching within specific directions or feature dimension groups. A single token, such as ``girl,'' may naturally form modification relationships with multiple appropriate adjectives (e.g., ``beautiful,'' ``young''). Consequently, the computational result tends to exhibit a diffusion effect, which we define as \textbf{DODM (Diffusion of Deep-Matching)}.

\vspace{0.5em}
The manifestation of DODM differs fundamentally between the two modes:
\begin{itemize}
    \item In the QKV mode, the DODM phenomenon occurs within the $V$ (Value) token group.
    \item In the QV mode, DODM occurs within the $Q$ (Query) component.
\end{itemize}

\begin{figure}[h]
    \centering
    \includegraphics[width=0.9\linewidth]{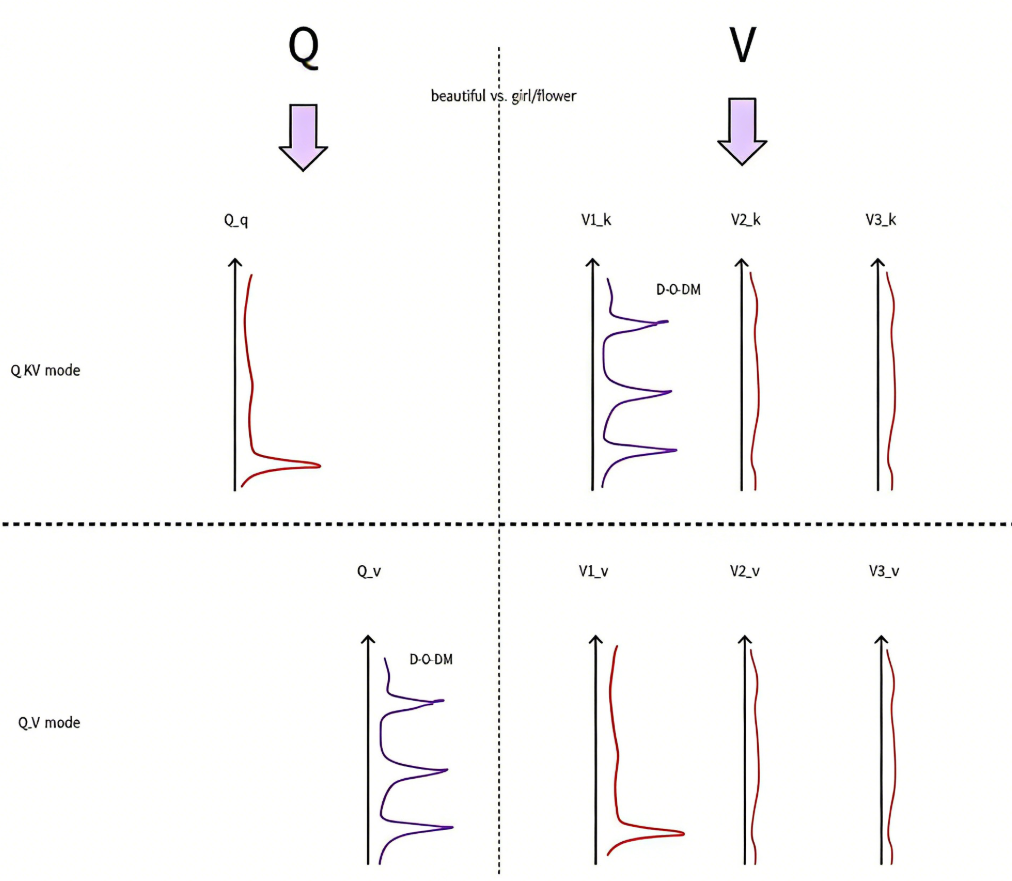}
    \caption{Diffusion of Deep-Matching}
    \label{fig:placeholder}
\end{figure}

Based on the Attention formulation, the process involves accumulating the dot-product results of $Q \cdot K$ and subsequently applying a \textit{softmax} operation across the scores of different $V$ vectors. In the  QKV mode, the DODM phenomenon allows the target matching words to be further reinforced and highlighted within the feature-dimensional context of the Query. 

\vspace{0.5em}
Conversely, in the  QV mode, DODM causes the query direction to exhibit a certain degree of \textit{divergence effect}. In a sense, the search intent becomes diffused rather than concentrated. We hypothesize that this structural difference—where $QKV$ concentrates attention while $QV$ inherently scatters the query focus—is the primary reason for the superior performance of the standard $QKV$ mechanism.

\section{MQA/GQA/MLA}

In recent advancements, KV-Shared architectures such as \textbf{GQA (Grouped-Query Attention)} \cite{ainslie2023gqa}and \textbf{MLA (Multi-head Latent Attention)} \cite{deepseek2024v2}have achieved significant success by drastically reducing the memory footprint of the KV-cache. We now seek to explain the underlying principles and essence of these mechanisms through our theoretical framework.

\subsection*{V-Shared}
To simplify the analysis, we introduce a \textit{minimal model hypothesis}: suppose each Token/Embedding possesses $N=6$ semantic or syntactic expression directions (e.g., verb, adjective, adverb, noun, etc.), denoted as $\{a, b, c, d, e, f\}$. The potential modifier-relationship combinations total $N \times N = 36$. For a model with 6 layers, this requires an average of 6 heads per layer. The core optimization problem is how to organize these heads to minimize the KV-cache.
\vspace{0.5em}

In each layer, we can consolidate the $V$ vectors. For instance:
\begin{itemize}
    \item \textbf{Scenario 1}: A single $V$ is shared by 6 Queries: $\{a, b, c, d, e, f\} \rightarrow V_a$.
    \item \textbf{Scenario 2}: Two $Vs$ are shared, each corresponding to 3 Queries: $\{a, b, c\} \rightarrow V_a$ and $\{d, e, f\} \rightarrow V_b$.
\end{itemize}
Compared to unconstrained matching, reusing $V$ within each layer significantly reduces the cache.

\subsection*{Variant of QV}
While the $QV$ mode naturally accommodates the above logic, a contradiction arises in the standard $QKV$ mode. According to our derivation, while $V_v$ (the value content) can be shared, $Q_q$ and $V_k$ must maintain a one-to-one correspondence for precise matching. It is logically inconsistent for multiple Headers in the same layer to share a single $V_k$ within the $QKV$ framework. 

\vspace{0.5em}
How, then, do we explain the success of MQA and GQA? We posit that there is only one possibility: the essence of MQA/GQA has shifted from the original $Q_q \cdot V_k V_v$ form toward a \textbf{variant of the $QV$ mode}, specifically:
\begin{equation}
    Q_v \cdot K_v V_v
\end{equation}
In this context, $K_v$ can be viewed as a version of $V$ injected with positional encodings. Its primary—and perhaps only—functional significance is the computation of positional relationships (assuming $D_k = D_v$).
\begin{figure}
    \centering
    \includegraphics[width=0.6\linewidth]{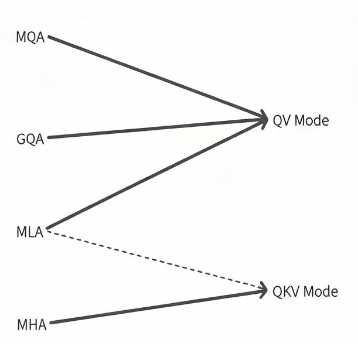}
    \caption{QV Mode vs QKV Mode}
    \label{fig:placeholder}
\end{figure}

\subsection*{MLA: The Optimized Evolution}
In industry practice, GQA often uses a fixed group size (e.g., 2, 4, or 8). However, this is suboptimal; different semantic features require varying numbers of Attention headers. MLA addresses this by applying a compression/extraction layer to $K$ and $V$, dynamically adapting to each Header. By employing this method, MLA outperforms GQA in terms of model accuracy, which comes at the cost of increased computational complexity.

\vspace{0.5em}
Because MLA performs deep compression on both $K$ and $V$, it cannot conform to the rigid $Q_q \cdot V_k V_v$ paradigm. Instead, MLA should be understood as a \textit{soft}, highly optimized version of GQA. Fundamentally, it remains a manifestation of the $QV$ mode, where the compressed latent space facilitates a more flexible semantic-matching-to-value-retrieval mapping.

\subsection*{Future Optimization Directions}

Based on the preceding analysis, we can derive several potential optimization trajectories for GQA and MLA:
\begin{itemize}
    \item \textbf{The GQA/MLA-QvVv Architecture}: If we adopt relative positional calculation methods such as \textbf{AGF}, \textbf{T5}, or \textbf{ALiBi}\cite{press2021train}, the $K$ vector can be entirely discarded. We designate this model as \textbf{GQA/MLA-QvVv} (assuming $D_k = D_v$). In this configuration, the traditional dependency on an explicit Key is eliminated, further streamlining the Attention mechanism while maintaining semantic integrity via the $QV$ logic.

    \item \textbf{Transition from KV-Shared to V-Shared Mode}: A strategic shift can be made from a \textit{KV-shared} to a \textbf{\textit{V-shared}} paradigm. Under this mode, the $V$ component remains maximally shared across heads to preserve memory efficiency. However, we retain a unique, one-to-one $K$ for each $Q$, transitioning the model back toward the original structural logic: $(Q_k \cdot V_k) V_v$. Although $K$ will occupy additional storage space in the KV-cache, this approach provides a richer representation for matching, which theoretically leads to higher model accuracy.

    \item \textbf{Optimization of Compression Ratios in MLA}: In current  MLA implementations, the compression of $K$ and $V$ is highly aggressive. We hypothesize that lowering the compression ratio of $K$—for instance, aligning it more closely with the compression level of $Q$—could preserve more global features. This would enable the model to better express the complex semantics inherent in the standard $(Q_q \cdot V_k) V_v$ mode, potentially yielding significant gains in predictive precision.
\end{itemize}

\section{Key-After-Value Mode (QV-Ka)}

Under the standard $(Q_q \cdot V_k) V_v$ paradigm, particularly in KV-Cache scenarios, $V$ theoretically offers significant potential for compression and reuse. However, since $K$ must maintain a one-to-one correspondence with $Q$, a critical question arises: is there further room for compressing $K$?

\vspace{0.5em}
According to our derived definition of the $QKV$ mechanism, a strong intrinsic correlation exists between $K$ and $V$. Specifically, $K$ can be interpreted as the result of a \textit{Deep-Matching} operation performed on the features of $V$, guided by the token-context. This relationship can be expressed as:
\begin{equation}
    K = \text{DM}(V, \text{Context})
\end{equation}
Consequently, it is theoretically feasible to reuse the information contained within $V$ for the computation of $K$.

\begin{figure}[h]
    \centering
    \includegraphics[width=0.5\linewidth]{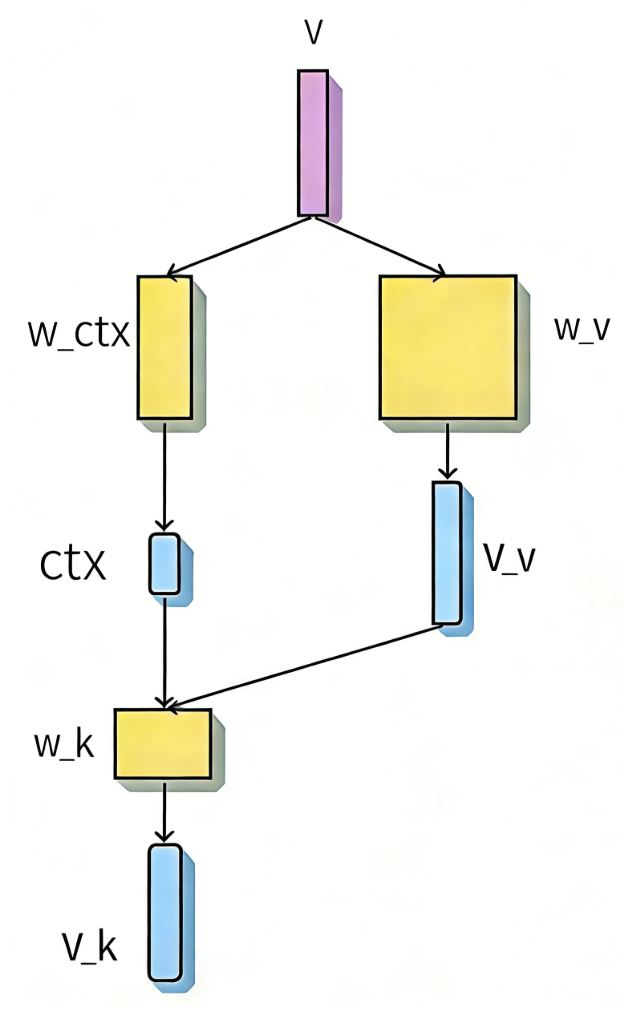}
    \caption{QV-Ka  Process}
    \label{fig:placeholder}
\end{figure}

The traditional Multi-Head Attention logic is defined as:
\begin{equation}
\begin{aligned}
\operatorname{MultiHead}(Q, K, V) &= \operatorname{Concat}(\mathrm{head}_1, \ldots, \mathrm{head}_h) W^O \\
\text{where } \mathrm{head}_i &= \operatorname{Attention}(Q W_i^Q, K W_i^K, V W_i^V)
\end{aligned}
\end{equation}

In our proposed model, the computational logic for each individual header is redefined as follows:
\begin{equation}
\begin{aligned}
    Q_i &= QW_i^Q \\
    V_i &= VW_i^V \\
    G &= KW_{ctx} \\
    K_i &= [G; V_i] W_i^K \\
    \mathrm{head}_i &= \operatorname{Attention}(Q_i, K_i, V_i)
\end{aligned}
\end{equation}

We designate this architecture as the \textbf{QV-Ka} (Key-after-value\&ctx) mode.
\vspace{0.5em}

In our evaluation, we maintained the default hyperparameter configuration: $L=3$, $d_{model} = 1024$, $h = 16$, and $d_{head} = d_{model}/h = 64$. We tested the \textbf{QV-Ka} mode by setting the dimension of the context vector ($d_{ctx}$) to $1\times$ and $2\times$ the size of $d_{head}$, respectively. The comparative results are summarized in Table~\ref{tab:qvka_results}.

\vspace{0.5em}

\begin{table}[h]
    \centering
    \caption{Performance Comparison under AGF and PCM-V Frameworks}
    \label{tab:qvka_results}
    \begin{tabular}{llc}
        \hline
        \textbf{Mode} & \textbf{Crafts } & \textbf{Valid Accuracy (\%)} \\
        \hline
        QV             & Default (Sinusoidal PE)         & 70.0756 \\
        QKV            & Default (Sinusoidal PE)         & 70.5911 \\
        \hline
        QV             & AGF + PCM-V                    & 70.5188 \\
        QKV            & AGF + PCM-V                    & 70.7305 \\
        QV-Ka ($d_{ctx} = d_{head}$) & AGF + PCM-V      & 70.4998 \\
        QV-Ka ($d_{ctx} = 2 d_{head}$) & AGF + PCM-V    & 70.6919 \\
        \hline
    \end{tabular}
\end{table}

\begin{figure}[h]
    \centering
    \includegraphics[width=0.9\linewidth]{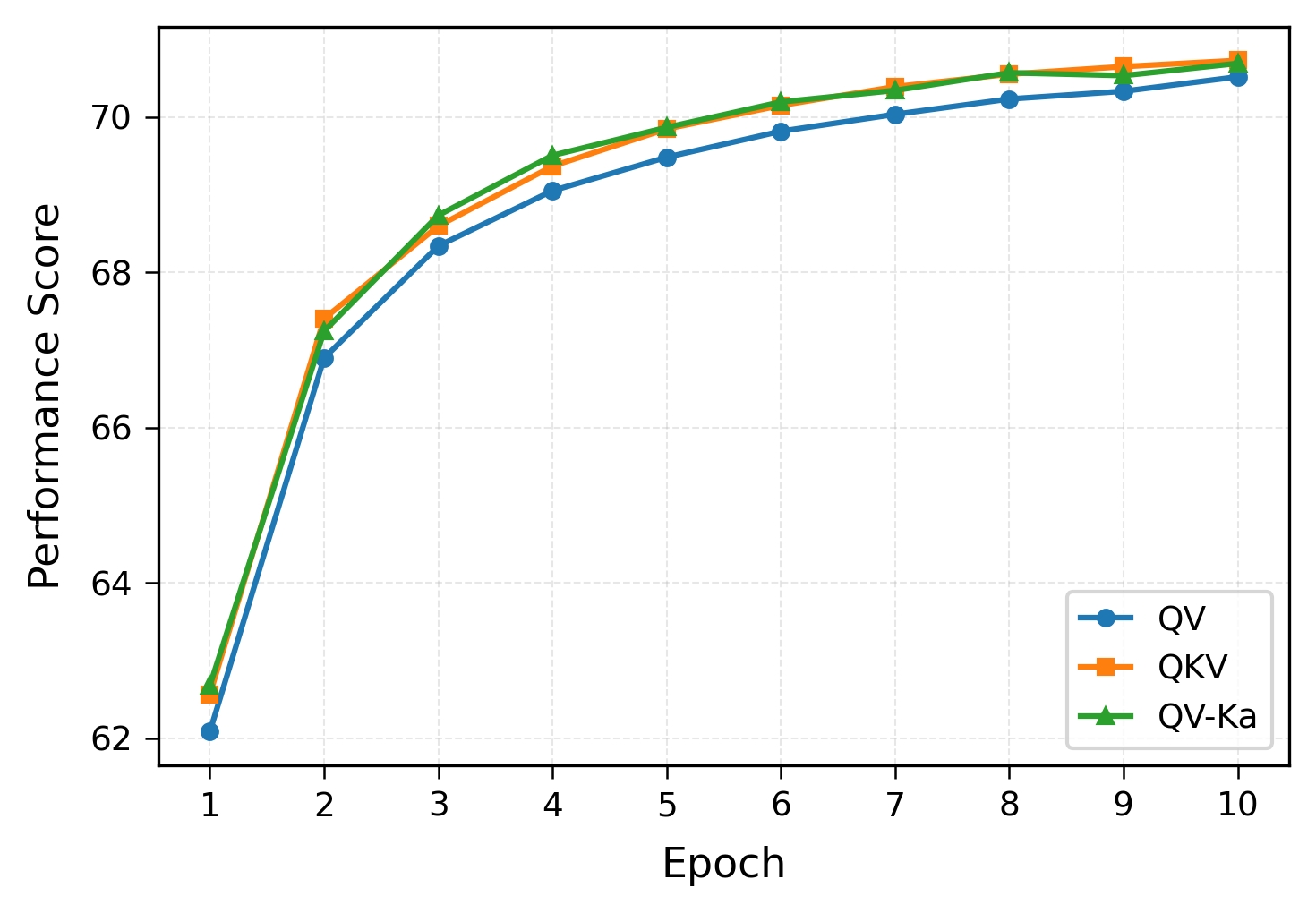}
    \caption{Comparision of QV/QKV/QV-Ka}
    \label{fig:placeholder}
\end{figure}

As illustrated by the performance curves under the AGF/PCM-V framework, the \textbf{QV-Ka ($2 \cdot d_{head}$)} mode achieves results comparable to the original \textbf{QKV} mode. Notably, this is accomplished with a significantly reduced parameter count and computational overhead. 

\vspace{0.5em}

Furthermore, during the initial stages of training, the \textbf{QV-Ka} mode exhibits slightly superior performance compared to the standard \textbf{QKV}. This observation aligns with our theoretical derivation of the $Q, K, V$ factors: the \textbf{QV-Ka} mode adheres more closely to the intrinsic logic of the problem, whereas the standard \textbf{QKV} mechanism can be viewed as a redundant, approximate implementation.

\section{Conclusion}

Departing from traditional experiment-driven or purely empirical approaches to model architecture design and optimization, this paper first establishes a theoretical and interpretable framework to hypothesize and demonstrate the intrinsic logic of the $QKV$ Attention mechanism. Through this logical derivation, we propose the \textbf{QV} mode as a fundamental abstraction. 
\vspace{0.5em}

By analyzing and synthesizing contemporary architectures such as  MQA, GQA, and MLA, we conclude that these models are essentially variants of the $QV$ paradigm. Building upon this insight, we provide strategic directions for future architectural optimizations and introduce the \textbf{QV-Ka} optimization mode. 

\vspace{0.5em}
Our experimental results provide robust empirical evidence supporting the validity of our theoretical conjectures. The \textbf{QV} and \textbf{QV-Ka} modes proposed in this study contribute directly to the enhancement of model efficiency and performance. Furthermore, we believe that our in-depth analysis of the underlying logic of $QKV$ Attention offers a valuable theoretical reference for the ongoing evolution of Large Language Model (LLM) architectures.

\section{Limitations}

Due to constraints in the experimental environment, the data scale, model parameters, and the number of layers utilized in this study remain relatively modest. While our findings provide strong theoretical and empirical support for the proposed mechanisms, their performance and scalability in production-grade Large Language Models (LLMs) with significantly larger parameter counts require further rigorous testing and validation.

\printbibliography[resetnumbers=True]

@inproceedings{vaswani2017attention,
  title={Attention is all you need},
  author={Vaswani, Ashish and Shazeer, Noam and Parmar, Niki and Uszkoreit, Jakob and Jones, Llion and Gomez, Aidan N and Kaiser, {\L}ukasz and Polosukhin, Illia},
  booktitle={Advances in neural information processing systems},
  pages={5998--6008},
  year={2017}
}

@article{zhang2026agf,
  title={Attention's Gravitational Field: A Power-Law Interpretation of Positional Correlation},
  author={Zhang, Edward},
  journal={arXiv preprint arXiv:2603.04805},
  year={2026}
}

@article{bhojanapalli2020low,
  title={Low-rank bottleneck in multi-head attention models},
  author={Bhojanapalli, Srinadh and Chakrabarti, Darshan and Papailiopoulos, Dimitris and Siddhardh, Rama Krishna and Srebro, Nathan},
  journal={arXiv preprint arXiv:2002.07028},
  year={2020}
}

@article{deepseek2024v2,
  title={DeepSeek-V2: A Strong, Economical, and Efficient Mixture-of-Experts Language Model},
  author={DeepSeek-AI and codes, others},
  journal={arXiv preprint arXiv:2405.04434},
  year={2024}
}

@article{shazeer2019mqa,
  title={Fast Transformer Decoding: One Write-Head is All You Need},
  author={Shazeer, Noam},
  journal={arXiv preprint arXiv:1911.02150},
  year={2019}
}

@article{ainslie2023gqa,
  title={GQA: Training Generalized Multi-Query Transformer Models from Multi-Head Checkpoints},
  author={Ainslie, Joshua and Lee-Thorp, James and de Jong, Michiel and Zemlyanskiy, Yury and Fedus, Federico and Muennighoff, Niklas and others},
  journal={arXiv preprint arXiv:2305.13245},
  year={2023}
}

@article{press2021train,
  title={Train Short, Test Long: Attention with Linear Biases Enables Input Length Extrapolation},
  author={Press, Ofir and Smith, Noah A and Lewis, Mike},
  journal={arXiv preprint arXiv:2108.12409},
  year={2021}
}

\end{document}